# Dual-stream Spatio-Temporal GCN-Transformer Network for 3D Human Pose Estimation

Jiawen Duan[a], Jian Xiang[a,*], Zhiqiang Li[a],Linlin Xue[a]

[a]*School of Information and Electronic Engineering, Zhejiang University of Science and Technology, Liuxia, Hangzhou 310023, China*

**Abstract**

3D human pose estimation is a classic and important research direction in the field of computer vision. In recent years, Transformer-based methods have made significant progress in lifting 2D to 3D human pose estimation. However, these methods primarily focus on modeling global temporal and spatial relationships, neglecting local skeletal relationships and the information interaction between different channels. Therefore, we have proposed a novel method—the Dual-stream Spatio-temporal GCN-Transformer Network (MixTGFormer). This method models the spatial and temporal relationships of human skeletons simultaneously through two parallel channels, achieving effective fusion of global and local features. The core of MixTGFormer is composed of stacked Mixformers. Specifically, the Mixformer includes the Mixformer Block and the Squeeze-and-Excitation Layer ( SE Layer). It first extracts and fuses various information of human skeletons through two parallel Mixformer Blocks with different modes. Then, it further supplements the fused information through the SE Layer. The Mixformer Block integrates Graph Convolutional Networks (GCN) into the Transformer, enhancing both local and global information utilization. Additionally, we further implement its temporal and spatial forms to extract both spatial and temporal relationships. We extensively evaluated our model on two benchmark datasets (Human3.6M and MPI-INF-3DHP). The experimental results showed that, compared to other methods, our MixTGFormer achieved state-of-the-art results, with P1 errors of 37.6mm and 15.7mm on these datasets, respectively.

*Keywords:* 3D human pose estimation; transformer; graph convolution

## 1. Introduction

3D human pose estimation is a hot topic in current human pose estimation research, aiming to accurately estimate the 3D coordinates of human keypoints from images or videos. Research in this field not only has theoretical significance but also shows great potential in various practical applications, such as motion analysis [1], virtual reality [2], augmented reality [3], and activity recognition [4,5].

* Corresponding author.
*E-mail address:* xiangjian@zust.edu.cn.

Current mainstream research methods for 3D human pose estimation include direct estimation [6,7] and 2D-to-3D lifting [8,9]. With the widespread application of 2D human pose detectors [10,11,12], the 2D-to-3D lifting approach now dominates 3D human pose estimation. From a technical perspective, 3D human pose estimation can be further divided into multi-view methods [13,14,15] and monocular methods [8,16]. Due to the widespread use of monocular RGB cameras in real-world scenarios, monocular methods have become the mainstream of current research. However, the depth uncertainty of monocular methods makes 3D pose estimation highly challenging. Since human joints are composed of local spatial and temporal dependencies, the spatial and temporal information carried by human motion can be utilized to address these challenges and improve the accuracy of 3D pose estimation [17,18,19].

For the 2D-to-3D pose estimation problem, two mainstream advanced models have emerged in recent years: Transformer-based models [20] and Graph Convolutional Network (GCN)-based models [21]. Transformers, initially successful in the field of natural language processing (NLP) [22], have also been applied to computer vision tasks, including human pose estimation. They excel at capturing long-range dependencies and have a powerful self-attention mechanism, which aligns well with the discrete nature of human joint representations and the long-range temporal dependency modeling required in skeleton sequences. Many studies have focused on this, such as Poseformer [8] and MixSTE [23]. Although Transformer-based human pose estimation has achieved good results, the global attention mechanism[24] it employs lacks sufficient attention to local spatial node relationships, which leads to the neglect of relationships between local spatial nodes. Additionally, since 2D pose sequences are flattened and input into the model, it is difficult to intuitively design the model based on the pose structure for tracing back local joint features. To address this issue, some researchers have begun to explore the use of Graph Convolutional Networks, which is a deep learning method based on graph-structured data that learns node representations by aggregating features from local neighborhoods. GCNs are adept at handling local dependencies, and since human skeletons can be represented as graph-structured data, GCN models can explicitly preserve the structure of 2D and 3D human poses during the convolutional propagation process. However, GCNs are not a perfect solution either; their limited number of layers restricts their ability to perceive long-range and global information. Therefore, combining GCN and Transformer to establish a unified architecture is highly effective, as it allows the network to simultaneously capture local and global dependencies.

To this end, we propose a novel method for 3D human pose estimation, called MixTGFormer, which can effectively handles both global and local information in the spatio-temporal dimension. Compared to existing fusion models, the core of MixTGFormer lies in its novel fusion backbone network, Mixformer. The core of Mixformer is the Mixformer Block, which effectively combines Transformer and GCN. Specifically, it adaptively fuses the features of Transformer and GCN, enabling the module to effectively integrate local skeletal relationships and global spatial relationships. Furthermore, we designed two forms of this module: the Spatial Mixformer Block and theTemporal Mixformer Block. This not only balances the local and comprehensive representation of human poses but also achieves simultaneous consideration of temporal and spatial dimensions. Additionally, we introduced the Squeeze-and-Excitation module (SE Layer) [25] into Mixformer. This module explicitly models the interdependencies between convolutional feature channels, further enhancing the model's representational capability. Through these efforts, we have improved the model's comprehensive feature extraction ability and increased the accuracy of 3D human pose estimation. In summary, our main contributions are as follows:

1. We propose a novel module, Mixformer Block, with both temporal and spatial forms. They aggregate the features of Transformer and GCN in the spatio-temporal dimension in a simple and effective manner.

2. We introduce the Squeeze-and-Excitation module and combine it with the Mixformer Block to form Mixformer, which further improves the model's performance by learning dependencies between different channels.
3. The MixTGFormer, built on Mixformer, outperforms other state-of-the-art methods on the Human3.6M and MPI-INF-3DHP datasets, achieving the best performance.

## 2. Related Work

### *2.1. 3D Human Pose Estimation*

3D human pose estimation is a classic and important problem in the field of computer vision, with decades of research history [26]. In the early stages [27,28,29], this work relied almost entirely on handcrafted features and geometric constraints as means to predict 3D human poses. With the rapid development of deep learning, deep learning has now become the primary method for 3D human pose estimation [30]. This problem can be classified from different perspectives, such as based on input data and estimation methods.

From the perspective of input data, the input data can be divided into multi-view and monocular views. While multi-view methods can provide richer spatial information, they require the simultaneous use of multiple cameras from different angles, which is not highly feasible in practical application scenarios. Monocular methods, although lacking depth information, are simpler in data collection, have lower hardware costs, and are easier to deploy and use. Additionally, the rapid development of computer vision technology has largely compensated for the depth issue, making monocular methods the mainstream approach. This study also uses monocular input.

From the perspective of estimation methods, 3D pose estimation can be divided into direct estimation and 2D-to-3D lifting. Direct estimation directly estimates 3D poses from images using convolutional networks [31], characterized by simplicity but lower accuracy. For example, Pavlakos et al. [7] used voxel likelihood to represent the confidence of joint positions in 3D space and inferred joint details through 3D heatmaps, but the model showed sensitivity to irrelevant factors. The 2D-to-3D lifting method first uses a 2D human pose detector to extract 2D poses, and then an independent method lifts the estimated 2D human poses to 3D human poses. This method relies on effective 2D pose detectors, and researchers mainly focus on the lifting from 2D to 3D poses. This strategy typically yields higher data accuracy. The methods used in the pose lifting stage include Temporal Convolutional Networks (TCN) [32,33], GCN, and Transformer. Currently, the most commonly used methods are based on Transformer and GCN. Additionally, some researchers have recently mixed these methods and achieved excellent results. In this paper, our work is also based on the 2D-to-3D lifting method, with mixed improvements using Transformer and GCN.

### *2.2. Transformer-based Methods*

Transformer was first proposed by Vaswani et al. [20] and demonstrated outstanding performance in natural language processing (NLP). It later entered the field of 3D human pose estimation and also achieved good results. Poseformer [8] was the first work to completely use Transformer as the backbone network in 3D human pose estimation. It achieved prediction by modeling spatial and temporal information, significantly outperforming previous CNN-based methods. PoseformerV2 [18] expanded the receptive field by utilizing compact representations of lengthy skeleton sequences in the frequency domain and improved robustness to sudden movements in noisy data. The proposed method effectively fused temporal and frequency domain features. MHFormer [34] incorporated multi-hypothesis spatio-temporal feature hierarchies into the model, independently and mutually processing multiple hypothesis information of body joints in an end-to-end manner, and averaging the target 3D poses. P-STMO [35] proposed masked pose modeling, applying masked joint modeling to 3D human pose estimation through self-supervised learning. STCFormer [36] partitioned the input joint features into two partitions, separately performing spatial and temporal attention, and used Multi-Head Self-Attention (MHSA) to encapsulate spatial and temporal context in parallel. MotionBERT [19] proposed a unified perspective, learning general representations of human motion from large-scale, diverse data, and then completing various human-centered downstream video tasks in a unified paradigm. These methods use different approaches to combine the temporal and spatial relationships of human skeleton points from different perspectives, but the aggregation and collaborative use of these key information is still insufficient, which affects the further approximation of real human poses.

### *2.3. GCN-based Methods*

Due to their powerful dynamic relationship capture capabilities, Graph Convolutional Networks are widely used in skeleton-based action recognition. Due to their high computational efficiency and similar task types, they have also been extensively applied in 3D human pose estimation in recent years. SemGCN [37] introduced a semantic graph convolutional network, which uses a stacked structure as a non-local module. This architectural choice helps learn the weights between adjacent nodes, thereby enhancing the connections between 2D joints. Graph Stacked Hourglass Networks [38] proposed a graph-stacked hourglass network model to learn human skeleton representations at different scales. GLA-GCN [39] proposed a global-local learning architecture that leverages global spatio-temporal representations and local joint representations in GCN-based models for pose estimation. Although GCN-based methods have a lighter memory load, there is still a certain gap in performance compared to Transformer-based methods.

### *2.4. Fusion Methods*

Hybrid models that combine the advantages of Transformer and Graph Convolution have recently attracted much attention, and these integrated methods have produced state-of-the-art results on many public datasets. GraFormer [40] replaced the multi-layer perceptron of Transformer with learnable GCN layers to form the GraAttention module, while ChebGConv [41] modeled implicit connection relationships between non-adjacent joints. DiffPose [42] interleaved GCN layers with self-attention layers as a diffusion model, which can capture spatial features between joints based on human skeletons. MotionAGFormer [43] combined the advantages of both, adaptively fusing features extracted from Transformer and Graph Convolutional Networks, achieving a comprehensive and balanced representation of human motion. However, although these methods have achieved excellent results, similar to GCN-based methods, they are strong in learning spatial information of a single pose but relatively weak in learning temporal correlations between different frames.

## 3. Method

### *3.1. Overall Architecture*

In this section, we will comprehensively introduce the proposed method, which we call the Two-stream Mixed Temporal and Spatial Transformer (MixTGFormer). The goal of MixTGFormer is to lift 2D keypoint sequences to corresponding 3D pose sequences. It effectively combines MHSA and GCN to achieve good fusion and extraction of spatio-temporal information from the input and then lifts the output. The overall architecture of the model is shown in Fig. 1. The input of the model is a 2D input sequence with confidence scores $X \in \mathbb{R}^{T\times J\times 3}$, where T and J represent the number of frames and joints, respectively. First, the input is projected into $d$-dimensional features $F^0 \in \mathbb{R}^{T\times J\times d}$, and then learnable spatial position encoding $P^{s}_{pos} \in \mathbb{R}^{1\times J\times d}$ is added. Subsequently, we use stacked Mixformers to compute $F^i \in \mathbb{R}^{T\times J\times d}(i = 1, \dots, N)$ to capture the underlying 3D structure of the skeleton sequence, where N represents the network depth. Finally, we use a linear layer with a tanh activation function to map $F^N$ to a higher dimension to compute the motion representation $M \in \mathbb{R}^{T\times J\times d'}$, and estimate the human 3D pose $\widehat{P} \in \mathbb{R}^{T\times J\times 3}$ through a regression head.

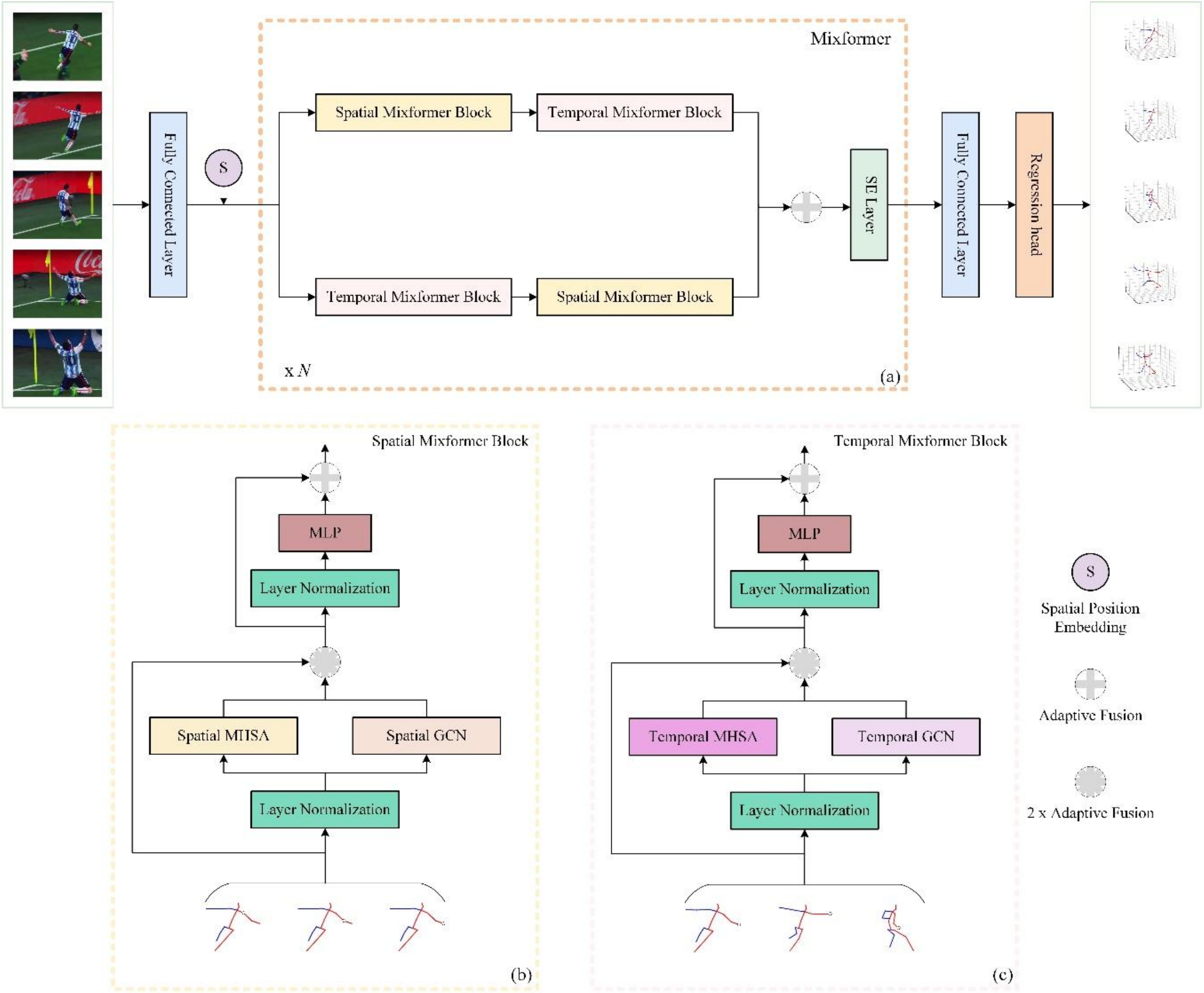

Fig. 1. Top: MixTGFormer model structure; (a) Overall architecture of Mixformer; (b) Spatial Mixformer Block; (c) Temporal Mixformer Block. The input tokens are the local joints of the human body and the frames of the pose sequence.

When performing the 2D-to-3D lifting task, losses may occur due to occlusion, detection failure, and errors. The loss terms include position loss ($L_{3D}$) and acceleration ($L_{\triangle A}$) loss, which are defined as:

$$L_{3D} = \sum_{t=1}^{T}\sum_{j=1}^{J} \| \widehat{P}_{t,j} - P_{t,j} \|, \quad L_{\triangle A} = \sum_{t=2}^{T}\sum_{j=1}^{J} \| \triangle \widehat{A}_{t,j} - \triangle A_{t,j} \|, \tag{1}$$

where $\widehat{A}_t = \widehat{P}_t - \widehat{P}_{t-1}$, $\triangle A_t = P_t - P_{t-1}$. Therefore, the total loss for 3D human pose estimation is:

$$L = L_{3D} + \lambda_{\triangle A} L_{\triangle A} + L_{\mathrm{2D}}, \tag{2}$$

where $\lambda_{\triangle A}$ is a constant coefficient used to balance position accuracy and motion smoothness, and $L_{\mathrm{2D}}$ is the loss generated by the 2D human pose detector predicting 2D poses.

Next, we first introduce the architectural design of the core part of MixTGFormer, Mixformer and then elaborate on the basic module composition of Mixformer.

*3.2. Mixformer*

Mixformer's core consists of two parts: the Spatio-Temporal Mixformer Block and the Squeeze-and-Excitation Layer (SE Layer, a self-attention mechanism). These two parts are sequentially connected, forming the overall structure (Fig. 1(a)). First, we constructed a dual-stream architecture containing two different forms of Mixformer Blocks for fusing the spatial and temporal information of the input. Specifically, we stacked the Mixformer Blocks in reverse order to form two parallel computational branches. Each branch has two forms of modules, temporal and spatial, so the spatio-temporal information is fused for the first time within the branch. Both branches model the keypoints by combining spatio-temporal information, but due to the different construction orders of the two branches, they have different emphases on spatio-temporal modeling. Subsequently, we fused the features extracted from the two streams using adaptive fusion, where the fusion is defined as:

$$F^i = \alpha_{ST}^i \circ F_{ST}^{i-1} + \alpha_{TS}^i \circ F_{TS}^{i-1}, \quad i \epsilon 1, \dots, N, \tag{3}$$

where $F^i$ represents the feature embedding at depth $i$, $\circ$ represents element-wise operations, $F_{ST}^{i-1}$, $F_{TS}^{i-1}$ represent the feature extraction performed by the Mixformer Block in the spatial-temporal and temporal-spatial order at depth $i-1$, N represents repeating the module fusion N times, and the adaptive fusion weights $\alpha_{TS}^i$,$\alpha_{TS}^i$ are defined as:

$$\alpha_{ST}^i, \alpha_{TS}^i = \mathrm{softmax}\left(W \cdot Concat\left(F_{ST}^{i-1}, F_{TS}^{i-1}\right)\right), \tag{4}$$

where W represents a learnable linear transformation.

After adaptive fusion, we use the SE layer for further computation to adaptively adjust the importance of each channel's features and improve the fusion of spatial and channel information, further enhancing the model's representational capability.

In summary, the Mixformer Block combines GCN and MHSA to achieve comprehensive modeling of local and global features of human skeletons, while the SE Layer further enhances the model's ability to model dependencies between channels. The combination of these two enables MixTGFormer to simultaneously capture global and local information of human poses in the spatial and temporal dimensions, significantly improving the accuracy of 3D human pose estimation.

*3.2.1. Mixformer Block*

The Mixformer Block is the core part of the model. To efficiently capture the global and local dependencies of human poses in different spatio-temporal dimensions, we combined GCN and multi-head attention mechanisms to form a novel backbone network to enhance the model's feature extraction and expressive capabilities. The Mixformer Block is designed in two modes: Spatial Mixformer Block and Temporal Mixformer Block, which mainly handle spatial information and temporal information, respectively. Their specific structures are shown in Fig. 1(b) and Fig. 1(c).

**Spatial Mixformer Block.** This module adopts a parallel processing structure of Spatial Multi-Head Self-Attention (S-MHSA) and Spatial GCN (S-GCN), treating individual joints as tokens to capture relationships between joints within a frame. Among them, S-MHSA is defined as:

$$\mathrm{S-MHSA}(Q_s, K_s, V_s) = Concat(head_i, \dots, head_h)W_s^{(O)},$$
$$head_i = softmax\left(Q_s^{(i)}\left(K_s^{(i)}\right)^T / \sqrt{d_k}\right) V_s^{(i)}, \quad (5)$$

where $W_s^{(O)}$ is the projection parameter matrix, $h$ is the number of parallel attention heads, and $d_k$ is the feature direction of $K_s$. To compute the query matrix $Q_s$, key matrix $K_s$, and value matrix $V_s$, we have

$$Q_s^i = F_s W_s^{(Q,i)}, K_s^i = F_s W_s^{(K,i)}, V_s^i = F_s W_s^{(V,i)}, \quad (6)$$

where $F_s \in \mathbb{R}^{BT \times J \times d}$ is the spatial feature, $W_s^{(Q,i)}, W_s^{(K,i)}, W_s^{(V,i)}$ are projection matrices, and $B$ is the batch size. S-GCN is defined as:

$$GCN\left(F^{(i)}\right) = \sigma\left(F^{(i)} + Norm\left(\widetilde{D} - 1/2\tilde{A}\widetilde{D} - 1/2F^{(i)}W_1 + F^{(i)}W_2\right)\right), \quad (7)$$

where $\tilde{A} = A + I_N$ represents the adjacency matrix with self-connections added, $I_N$ represents the identity matrix, $\widetilde{D}ii = \sum j \tilde{A}jj$ is defined as the sum of elements along the diagonal of $\tilde{A}$, $W_1$ and $W_2$ represent the trainable weight matrices specific to each layer, and $\sigma$ is the activation function.

Next, we adaptively fuse S-MHSA and S-GCN and add residual connections to capture the global and local spatial dependencies between joints. Then, the fused result is input into a multilayer perceptron (MLP), followed by LayerNorm and residual connection operations.

**Temporal Mixformer Block.** Similar to the Spatial Mixformer Block, this module has a similar structural flow, but the difference lies in the choice of MHSA and GCN. Specifically, the Temporal Mixformer Block uses Temporal Multi-Head Self-Attention (T-MHSA) and Temporal GCN (T-GCN) as components, so this module treats each frame as a token to capture relationships between consecutive frames. Among them, T-MHSA can be similarly expressed as:

$$\mathrm{T-MHSA}(Q_T, K_T, V_T) = Concat(head_i, \dots, head_h)W_T^{(O)},$$
$$head_i = softmax\left(Q_T^{(i)}\left(K_T^{(i)}\right)^T / \sqrt{d_k}\right) V_T^{(i)}, \quad (8)$$

where $Q_T$,$K_T$, and $V_T$ are computed similarly to formula (6). For T-GCN, it differs from S-GCN in their adjacency matrices and input features. Additionally, S-GCN uses $Sim\left(F_T^{t_i}, F_T^{t_j}\right) = (F_T^{t_i})^T F_T^{t_j}$ to calculate the similarity between individual joints in different time ranges. Subsequently, similar to the Spatial Mixformer Block, it also adaptively fuses S-MHSA and S-GCN and adds residual connections, followed by LayerNorm and residual connection operations.

### *3.2.2. Squeeze-and-Excitation Layer*

The SE Layer can be considered a self-attention mechanism, where the features of the convolutional layer are reweighted based on the average of all features in that layer, suppressing or emphasizing specific features by multiplying the corresponding features by appropriate scalars. Using it after the Mixformer Block further compensates for the deficiencies of GCN in global information modeling and MHSA in ignoring local structural dependencies, while reducing the error accumulation of human end joints.

The SE Layer uses the adaptively fused Mixformer Block as input. First, in the Squeeze stage, the number of time frames T and the number of joints J are pooled to compress the information in these two dimensions. The Squeeze process is defined as:

$$x_{squeezed} = 1/T \times J \sum_{t=1}^{T} \sum_{j=1}^{J} x_{b,t,j,c}\,, \quad (9)$$

where $x_{b,t,j,c}$ is the element of tensor x in the input dimension. Then, in the Excitation stage, we use a small MLP consisting of two fully connected layers (FC). The first fully connected layer (using ReLU activation function) performs dimensionality reduction, and then the second fully connected layer restores it. Finally, the Sigmoid activation function ensures that the obtained weights are between [0,1]. The Excitation stage can be expressed as:

$$x_{excited} = \sigma\left(FC2\left(ReLU\left(FC1(x_{squeezed})\right)\right)\right), \tag{10}$$

where $FC1$ and $FC2$ are two fully connected layers, and $\sigma$ is the Sigmoid activation function. Finally, in the Scale stage, the generated weights are applied to each channel of the input tensor to weight the input data. The final output is expressed as:

$$x' = x \cdot x_{excited}, \tag{11}$$

where $x$ is the input tensor, and $x_{excited}$ is the channel weighting coefficient obtained through the Excitation stage. The SE layer is shown in Fig. 2.

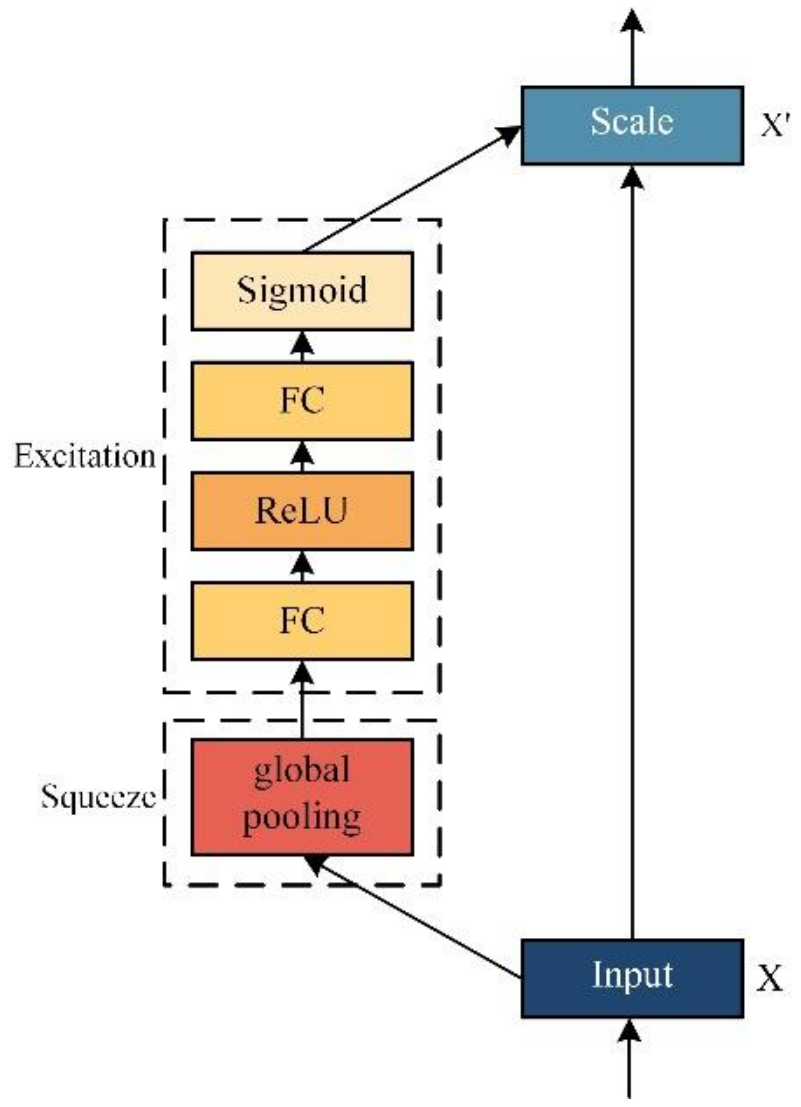

Fig. 2. Overall architecture of the Squeeze-and-Excitation Layer (SE Layer). The input X and output X' have the same shape.

## 4. Experiments

### *4.1. Datasets and Evaluation Metrics*

We comprehensively validated the proposed model (MixTGFormer) on two large-scale 3D human pose estimation datasets (Human3.6M [44] and MPI-INF-3DHP [45]).

The Human3.6M dataset is the most commonly used dataset in 3D human pose estimation. It includes 3.6 million video frames of 11 professional subjects performing 15 different daily activities, captured by 4 cameras from different perspectives. To ensure fair evaluation, we followed the evaluation method of most previous works, using the data of subjects 1, 5, 6, 7, and 8 for model training and the data of subjects 9 and 11 for testing. We selected two metrics to evaluate the model: MPJPE and P-MPJPE. MPJPE (called P1) calculates the mean per joint position error in millimeters between the estimated pose and the actual pose after aligning the root node (sacrum). P-MPJPE (called P2) requires the actual pose and the estimated pose to be aligned through rigid transformation to further calculate the loss. Fig. 3 shows the results of our model on the P1 metric.

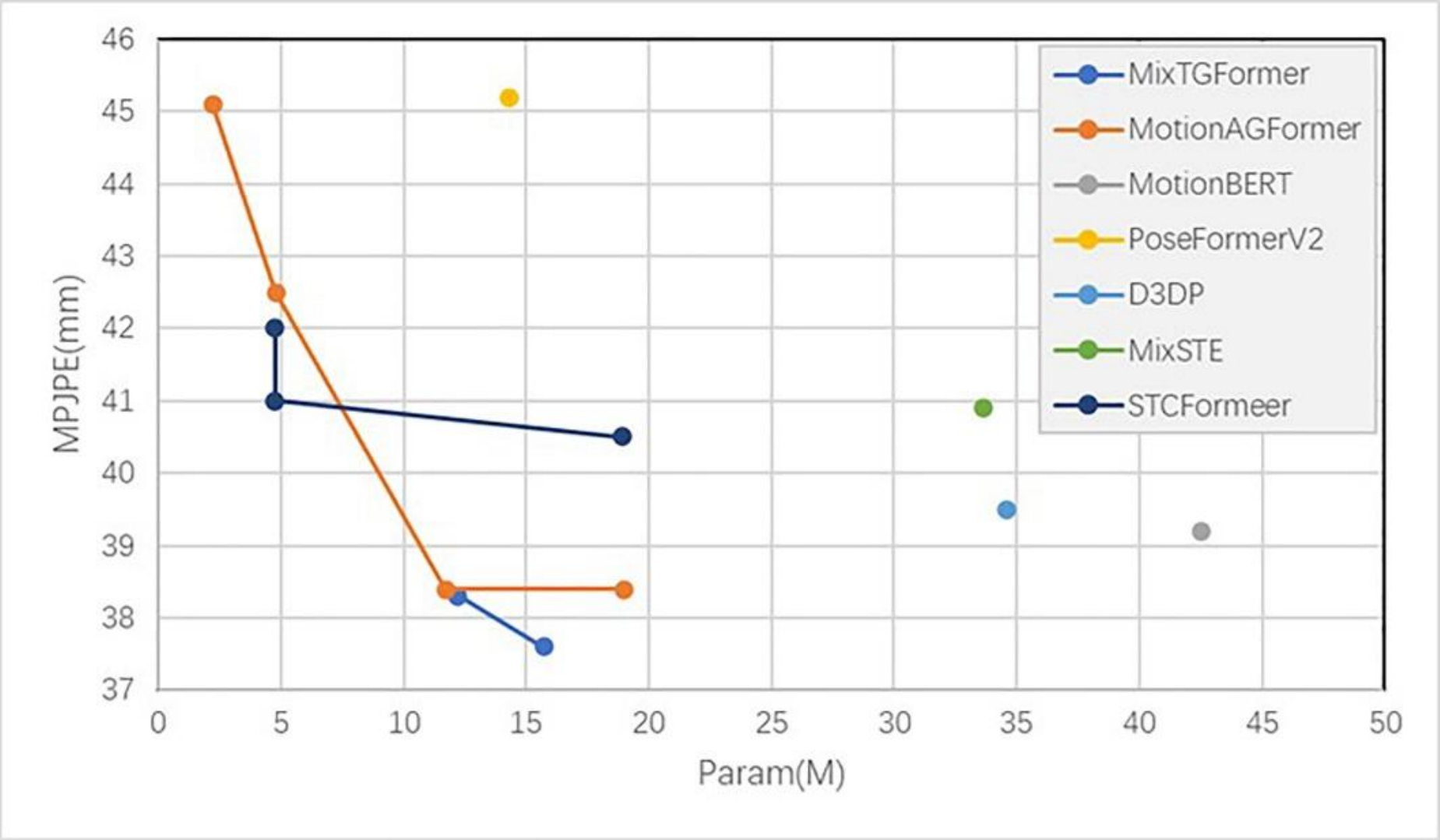

Fig. 3. Comparison with other 3D human pose estimation methods on the Human3.6M dataset. MPJPE represents the mean (per) joint position error (the lower the better), and Param represents the number of parameters.

The MPI-INF-3DHP dataset is another large-scale dataset commonly used in 3D human pose estimation, with three different settings: green screen, non-green screen, and outdoor environment. Following the evaluation method of previous works, we adopted MPJPE, the percentage of correct keypoints within 150mm (PCK), and the area under the curve (AUC) as evaluation metrics.

## *4.2. Implementation Details*

We constructed two models with different layer configurations: MixTGFormer-s and MixTGFormer, to adapt to different application requirements. The specific parameters are shown in Table 1.

Table 1. Details of the two versions of the MixTGFormer model. L: Number of layers. T: Number.

| Method | L | T | Params | MACs |
|---|---|---|---|---|
| MixTGFormer-s | 12 | 243 | 12.2M | 1.0G |
| MixTGFormer | 16 | 243 | 15.7M | 1.4G |

Our model is implemented based on the PyTorch framework, and all experiments were conducted on two NVIDIA RTX 4090 GPUs. Referring to previous research [43], we applied horizontal flip augmentation during training and testing. In training, we set 60 epochs of training and used the AdamW [46] optimizer with a weight decay of 0.01, a batch size of 4, and an initial learning rate of 5e-4. For the input, we followed the settings of previous research. For Human3.6M, we used the 2D detection results of Stacked Hourglass and the 2D ground truth provided by Human3.6M as input. For MPI-INF-3DHP, we used the 2D ground truth of MPI-INF-3DHP as input.

### 4.3. Results on Human3.6M

We compared MixTGFormer with other state-of-the-art methods on Human3.6M. For fair comparison, we only listed the results of models that were not pre-trained on additional data. To comprehensively evaluate the model's performance, we not only considered the model's performance but also compared the model parameters and computational complexity of each model. The experimental results are shown in Table 2, indicating that MixTGFormer achieved a P1 error of 37.6mm and a P2 error of 31.8mm, both of which are the best results to date. This result also demonstrates the advantage of our comprehensive fusion mechanism over the partial fusion used in previous methods in extracting global and local spatio-temporal relationships of human skeletons. Additionally, we achieved a very good performance in the real error $P1^{\dagger}$, which eliminates the influence of 2D detector errors, proving the overall effectiveness of the model.

Table 2. Quantitative comparison results on Human3.6M. T: Number of input frames; P1: MPJPE error (mm); P2: P-MPJPE error (mm); $P1^{\dagger}$: Real P1 error with 2D ground truth. The best and second-best results are shown in bold and underlined, respectively.

| Method | T | Param | MACs | P1↓ | P2↓ | $P1^{\dagger}$↓ |
|---|---|---|---|---|---|---|
| MHFormer[34] | 351 | 30.9M | 7.0G | 43.0 | 34.4 | 30.5 |
| MixSTE[23] | 243 | 33.6M | 139.0G | 40.9 | 32.6 | 21.6 |
| P-STMO[35] | 243 | 6.2M | 0.7G | 42.8 | 34.4 | 29.3 |
| STCFormer[36] | 243 | 4.7M | 19.6G | 41.0 | <u>32.0</u> | 21.3 |
| STCFormer-L[36] | 243 | 18.9M | 78.2G | 40.5 | **31.8** | - |
| PoseFormerV2[18] | 243 | 14.3M | 0.5G | 45.2 | 35.6 | - |
| GLA-GCN[39] | 243 | 1.3M | 1.5G | 44.4 | 34.8 | 21.0 |
| HDFormer[47] | 96 | 3.7M | 0.6G | 42.6 | 33.1 | 21.6 |
| HSTFormer[48] | 81 | 22.7M | 1.0G | 42.7 | 33.7 | 27.8 |
| MotionBERT[19] | 243 | 42.5M | 174.7G | 39.2 | 32.9 | 17.8 |
| MotionAGFormer-B[43] | 243 | 11.7M | 48.3G | 38.4 | 32.6 | 19.4 |
| MotionAGFormer-L[43] | 243 | 19.0M | 78.3G | 38.4 | 32.5 | 17.3 |
| DSTFormer[9] | 243 | 12.0M | 48.3G | <u>37.9</u> | 32.2 | **15.6** |
| MixTGFormer-s | 243 | 12.2M | 46.6G | 38.3 | 32.4 | 17.8 |
| MixTGFormer | 243 | 15.7M | 63.4G | **37.6** | **31.8** | <u>16.4</u> |

To provide a more intuitive comparison, we visualized some test results. Fig. 4 shows the 3D pose estimation results of MixTGFormer and some other models. For clarity, we used arrows to indicate the areas where our model showed significantly better pose estimation compared to other methods. Overall, compared to other methods, our method demonstrated more comprehensive and superior human body reconstruction results.

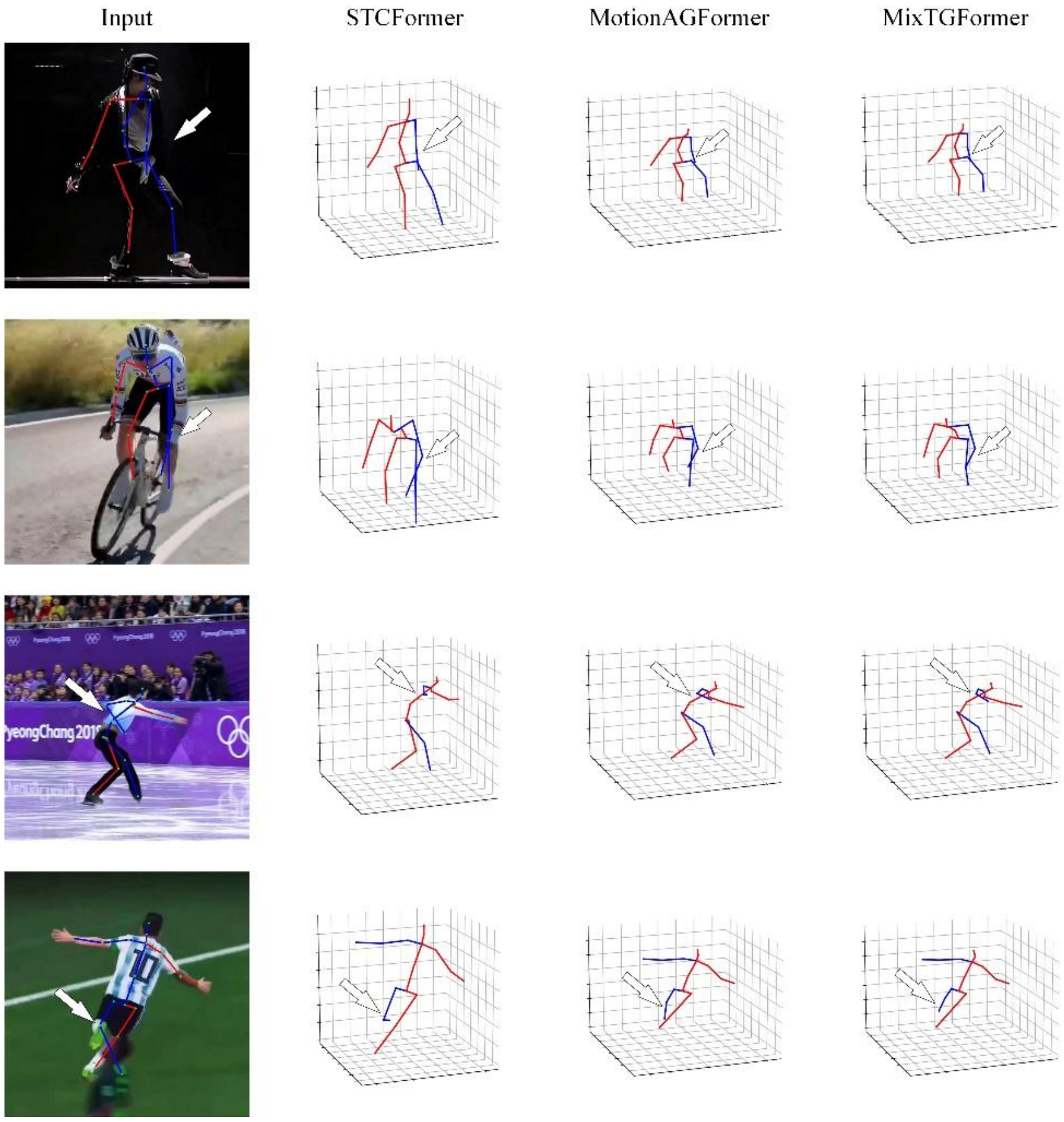

Fig. 4. Qualitative comparison of MixTGFormer with STCFormer and MotionAGFormer. Blue represents the predicted left side of the body, and red represents the predicted right side of the body. The arrows point to the estimation results that show significant differences.

### *4.4. Results on MPI-INF-3DHP*

On MPI-INF-3DHP, we compared MixTGFormer with other existing methods. Similar to MotionAGFormer, we used 81 frames as the input frame number. Table 3 shows the detailed results of the experiment. It can be seen that in the standard model, the AUC is 85.4%, and the MPJPE is 16.5mm, which is a significant improvement of 1.2% and a reduction of 1.7mm compared to the current best results. Additionally, on the PCK metric, our model is only 0.2% lower than the current best result. Furthermore, our small model also achieved very competitive results.

Table 3. Quantitative comparison results on MPI-INF-3DHP. T: Number of input frames. Detailed descriptions of the three evaluation metrics are provided in Section 4.1. The best and second-best results are shown in bold and underlined, respectively.

| Method | T | PCK↑ | AUC↑ | MPJPE↓ |
|---|---|---|---|---|
| MHFormer[34] | 9 | 93.8 | 63.3 | 58.0 |
| MixSTE[23] | 27 | 94.4 | 66.5 | 54.9 |
| P-STMO[35] | 81 | 97.9 | 75.8 | 32.2 |
| STCFormer[36] | 81 | **98.7** | 83.9 | 23.1 |
| PoseFormerV2[18] | 81 | 97.9 | 78.8 | 27.8 |
| GLA-GCN[39] | 81 | <u>98.5</u> | 79.1 | 27.7 |
| HSTFormer[48] | 81 | 97.3 | 71.5 | 41.4 |
| HDFormer[47] | 96 | **98.7** | 72.9 | 37.2 |
| MotionAGFormer[43] | 81 | 98.3 | 84.2 | <u>18.2</u> |
| MixTGFormer-s | 81 | 98.3 | <u>84.7</u> | 18.6 |
| MixTGFormer | 81 | <u>98.5</u> | **85.4** | **16.5** |

### *4.5. Ablation Studies*

In this section, we will gradually demonstrate some modifications made to MixTGFormer and the improvements brought by these modifications. All ablation experiments were evaluated on the Human3.6M dataset.

**Choice of Module Composition in Mixformer Block.** The composition of the Mixformer Block is the key to this method. We experimentally investigated the specific impact of different module compositions on the model's performance, and the results are listed in Table 4. "DoubleAttention" indicates that both branches in the Mixformer Block use MHSA, "DoubleGCN" indicates that both branches use GCN, and "Attention-GCN" indicates that one branch uses MHSA and the other uses GCN. Through comparison, we found that using different modules in the two branches reduced the P1 error by 1.0mm and 0.4mm, respectively, compared to other composition methods, proving the effectiveness of combining GCN's local spatial relationship capture capability with MHSA's global information extraction capability.

Table 4. Comparison of different module compositions in the Mixformer Block. The experiment was conducted under the spatial-temporal connection order of the Mixformer Block.

| Method Description | MPJPE(mm) | P-MPJPE(mm) |
|---|---|---|
| Double Attention | 38.5 | 32.3 |
| Double GCN | 39.1 | 32.9 |
| Attention-GCN | 38.1 | 32.3 |

**Choice of Connection Order in Two Mixformer Blocks.** Considering that different spatio-temporal orders may affect the model's performance, we further investigated the model's performance under different connection orders, and the results are shown in Table 5. We can see that when both branches use a single spatio-temporal module, the model's performance is the worst. When both parallel branches use the spatial-temporal connection, the P1 error is 38.1mm. When both branches are inverted to the temporal-spatial order, the error increases to 38.3mm. When the two branches use the temporal-spatial and spatial-temporal orders, respectively, the error decreases by 0.2mm. This verifies the necessity and effectiveness of our design of the two Mixformer Blocks and the cross-connection order.

Table 5. Comparison of different connection orders of the Spatial Mixformer Block and Temporal Mixformer Block. S: Spatial Mixformer Block. T: Temporal Mixformer Block.

| Method Description | MPJPE(mm) | P-MPJPE(mm) |
|---|---|---|
| Double S—>S | 38.7 | 32.6 |
| Double T—>T | 38.6 | 32.9 |
| Double S—>T | 38.1 | 32.3 |
| Double T—>S | 38.3 | 32.0 |
| S—>T and T—>S | 37.9 | 31.9 |

**Choice of Different Encoding Embeddings.** Similar to MotionAGFormer, we also explored the impact of different encoding embeddings on the model, and the experimental data are shown in Table 6. When we used temporal encoding embeddings, the model produced a P1 error of 38.9mm. When we switched to spatial encoding embeddings, the error decreased significantly by 1.0mm. Surprisingly, when we introduced both encoding embeddings into the model, the error decreased to the maximum.

Table 6. Comparison of models using different encoding embeddings.

| Temporal Embedding | Spatial Embedding | MPJPE(mm) | P-MPJPE(mm) |
|---|---|---|---|
| √ | | 38.9 | 32.6 |
| | √ | 37.9 | 31.9 |
| √ | √ | **39.2** | **32.8** |

**Choice of SE Layer Insertion Position.** In Table 7, we studied the impact of introducing the SE Layer and its different insertion positions on the model. Here, we tried three different insertion positions. When the SE Layer was inserted between the Mixformer Block and adaptive fusion or between two sequentially connected Mixformer Blocks, the performance was similar, both producing a P1 error of 38.1mm, which was a 0.2mm increase compared to the original model. However, when the insertion position was between adaptive fusion and the fully connected layer, the P1 error decreased by 0.3mm. This fully proves the SE Layer's compensation for GCN's insufficient global modeling and MHSA's neglect of local dependencies, confirming the effectiveness of the SE Layer in improving model performance.

Table 7. Comparison of different insertion positions of the SE Layer. a: No insertion; b: Inserted between the Mixformer Block and adaptive fusion; c: Inserted between two sequentially connected Mixformer Blocks; d: Inserted between adaptive fusion and the fully connected layer.

| Insertion position | MPJPE(mm) | P-MPJPE(mm) |
|---|---|---|
| a | 37.9 | 31.9 |
| b | 38.1 | 31.9 |
| c | 38.1 | 32.0 |
| d | 37.6 | 31.8 |

## 5. Conclusion

In this paper, we proposed MixTGFormer, a novel Transformer and GCN-based 3D human pose estimation model. It adopts a dual-stream fusion mechanism in both the backbone network and core components, which enhances the model's understanding and feature-capture capabilities of global and local spatio-temporal relationships of human skeletons. Additionally, we introduced the Squeeze-and-Excitation Layer to further enhance the model's ability to address the neglect of specific modeling features. Extensive experimental evaluations demonstrated the effectiveness of our method on Human3.6M and MPI-INF-3DHP, with results surpassing current state-of-the-art algorithms.

Although our model has achieved excellent performance on multiple benchmark datasets, there are still some potential directions for improvement worth exploring, including further optimizing the model structure to reduce computational complexity or expanding the model's application scenarios, such as real-time pose tracking and multi-view pose estimation.

## Funding

This work was supported by the Zhejiang Provincial Key Research and Development Project(No. 2025C02045).

## CRediT authorship contribution statement

**Jiawen Duan:** Writing – original draft, Validation, Visualization, Investigation. **Jian Xiang:** Writing – review and editing, Supervision, Methodology. **Zhiqiang Li:** Data curation, Formal analysis. **Linlin Xue:** Investigation, Methodology.

## Declaration of competing interest

The authors declare that they have no known competing financial interests or personal relationships that could have appeared to influence the work reported in this paper.

## Data availability

Data will be made available on request.